%% file: paper.tex
\documentclass[authoryear,preprint,12pt]{elsarticle}

\usepackage{amssymb}
\usepackage{hyperref}
\usepackage{url}
\input{math_commands.tex}

\usepackage{color}   
\usepackage{hyperref}
\hypersetup{
	colorlinks=true, 
	linktoc=all,     
	linkcolor=blue,  
	citecolor=red,
}

\usepackage{geometry}                

\usepackage{subfigure}
\usepackage{amssymb}
\usepackage{amsmath}
\usepackage{appendix}
\usepackage{tikz}
\usepackage{ulem}
\usepackage{tabularx}
\usepackage{wrapfig}

\usepackage{booktabs}       
\usepackage{multirow}
\usepackage{mathrsfs}
\usepackage{scrextend}
\usepackage{pdfpages}
\usepackage{diagbox}

\usetikzlibrary{positioning}
\usetikzlibrary{arrows}

\numberwithin{figure}{section}
\numberwithin{equation}{section}

\usepackage{algorithm}
\usepackage{algorithmic}
\usepackage{bm}
\usepackage{graphicx} 
\usepackage{subfigure}

\usepackage{stmaryrd}

\graphicspath{{figures}}

\begin{document}


\begin{frontmatter}



\title{Dilated convolution neural operator for multiscale partial differential equations}


\author[label1]{Bo Xu}

\author[label2]{Xinliang Liu}

\author[label3]{Lei Zhang}

\affiliation[label1]{organization={Shanghai Jiao Tong University},
            addressline={School of Mathematical Sciences}, 
            city={Shanghai},
            postcode={200240},
            country={China}}           
\affiliation[label2]{organization={King Abdullah University of Science and Technology},
            addressline={Computer, Electrical and Mathematical Science and Engineering Division, King Abdullah University of Science and Technology}, 
            city={Thuwal},
            postcode={23955}, 
            country={Saudi Arabia}}
\affiliation[label3]{organization={Shanghai Jiao Tong University},
            addressline={School of Mathematical Sciences, Institute of Natural Sciences, and MOE-LSC}, 
            city={Shanghai},
            postcode={200240},
            country={China}}
\begin{abstract}
    This paper introduces a data-driven operator learning method for multiscale partial differential equations, with a particular emphasis on preserving high-frequency information. Drawing inspiration from the representation of multiscale parameterized solutions as a combination of low-rank global bases (such as low-frequency Fourier modes) and localized bases over coarse patches (analogous to dilated convolution), we propose the Dilated Convolutional Neural Operator (DCNO). The DCNO architecture effectively captures both high-frequency and low-frequency features while maintaining a low computational cost through a combination of convolution and Fourier layers. We conduct experiments to evaluate the performance of DCNO on various datasets, including the multiscale elliptic equation, its inverse problem, Navier-Stokes equation, and Helmholtz equation. We show that DCNO strikes an optimal balance between accuracy and computational cost and offers a promising solution for multiscale operator learning.
\end{abstract}




\end{frontmatter}



\section{Introduction}
\label{sec:introduction}

In recent years, operator learning methods such as Fourier Neural Operator (FNO) \citep{li2020fourier}, Galerkin Transformer (GT) \citep{cao2021choose}, and Deep Operator Network (DeepONet) \citep{lu2021learning} have emerged as powerful tools for computing parameter-to-solution maps of partial differential equations (PDEs). In this paper, we focus on multiscale PDEs that encompass multiple temporal/spatial scales. These multiscale PDE models are widely prevalent in physics, engineering, and other disciplines, playing a crucial role in addressing complex practical problems such as reservoir modeling, atmosphere and ocean circulation, and high-frequency scattering.

A well-known challenge with neural networks is their tendency to prioritize learning low-frequency components before high frequencies—a phenomenon referred to as "spectral bias" or "frequency principle" \citep{rahaman2018spectral,xu2020frequency}. This presents challenges when applying neural networks to functions characterized by multiscale or high-frequency properties, adapting neural network architectures \citep{cai2019multiscale,WANG2021113938} have been proposed to capture high-frequency details . In the context of operator learning, existing methods such as FNO and GT have shown spectral bias when applied to multiscale PDEs, as observed in \citet{HANO}. To address this issue and recover high-frequency features, \citet{HANO} introduced an approach based on hierarchical attention and $H^1$ loss. However, despite providing improved accuracy, the method's high computational cost and implemational complexity to some extent counterbalances its strength.

In this paper, we present a novel method that strikes a balance among accuracy, computational cost, and the preservation of multiscale features. Our approach utilizes a carefully designed architecture that combines the strengths of dilated convolutions and Fourier layers. Dilated convolutions \citep{Holscheider_book_dilation}, also known as atrous convolutions, expand the kernel of a convolution layer in a convolutional neural network (CNN) by introducing gaps (holes) between the kernel elements. This technique allows for selectively skipping input values with specific step sizes, effectively covering a larger receptive field over the input feature map without introducing extra parameters or computational overhead. As a result, we can efficiently capture high-frequency local details. On the other hand, we leverage Fourier layers to capture the smooth global components of the data. DCNO achieves higher accuracy compared to existing models while maintaining lower computational costs by utilizing efficient implementations of both convolution and Fourier layers. This makes our approach well-suited for applications that require the preservation of multiscale features.

\begin{figure}[H]
    \centering
    \subfigure[\scriptsize multiscale trigonometric coefficient,]{\includegraphics[width=0.23\textwidth]{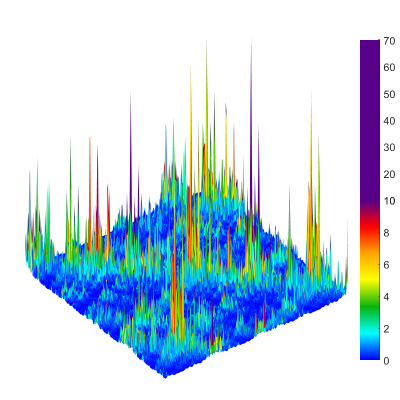}}
    \subfigure[\scriptsize slices of the derivatives $\frac{\partial u}{\partial y}$ at $x=0$,]{\includegraphics[width=0.23\textwidth]{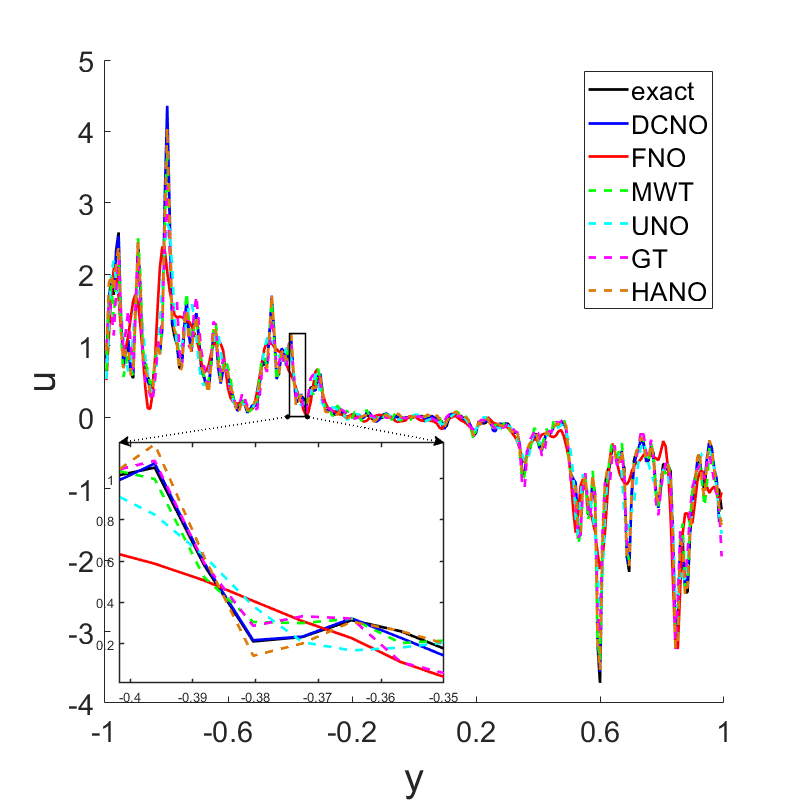}}
    \subfigure[\scriptsize (testing) low frequency error dynamics in spectral domain]{\includegraphics[width=0.23\textwidth, height=0.23\textwidth]{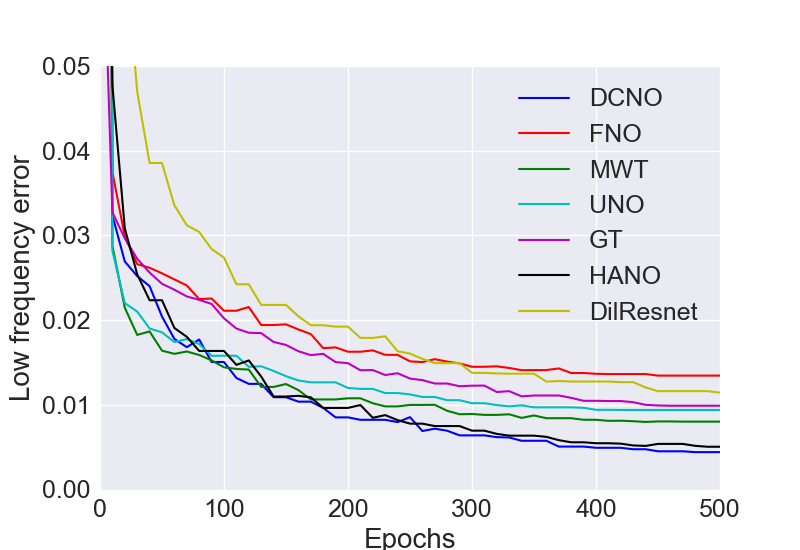}}
    \subfigure[\scriptsize (testing) high frequency error dynamics in spectral domain]{\includegraphics[width=0.23\textwidth, height=0.23\textwidth]{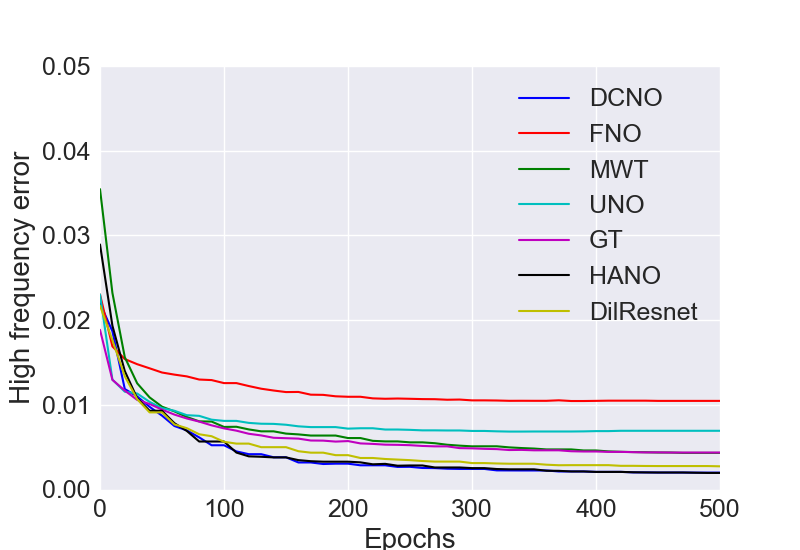}}
    \caption{We demonstrate the effectiveness of the DCNO scheme using a challenging multiscale trigonometric benchmark. The coefficient and corresponding solution derivative are presented in (a) and (b), respectively (refer to Section \ref{sec:mul}) for a detailed description. We observe that DCNO accurately captures the solution derivatives. In (c) and (d), we analyze the (testing) dynamics for high-frequency ($> 10\pi$) and low-frequency ($\leq 10\pi$) errors, respectively. It is evident that DCNO achieves outstanding performance in terms of both high-frequency and low-frequency errors.}
    \label{fig:gamblet_fig}
\end{figure}

\section{Background and Related Work}
\label{sec:background}
\subsection{Multiscale PDEs }
\label{sec:background:mspde}

\def\aspace{\mathcal{A}\left(D ; \mathbb{R}^{d_{a}}\right)}
\def\uspace{\mathcal{U}\left(D ; \mathbb{R}^{d_{u}}\right)}
\def\L{\mathcal{L}}
\def\B{\mathcal{B}}
\def\H{\mathcal{H}}

We briefly introduce some representative multiscale PDEs in this section. One notable example is the class of \emph{multiscale elliptic PDEs}, which involve coefficients varying rapidly and are often encountered in  heterogeneous and random media applications. For smooth coefficients, the coefficient to solution map can be effectively resolved by the FNO parameterization \citep{li2020fourier}. However, when dealing with multiscale/rough coefficients, the presence of fast oscillation, high contrast ratios, and non-separable scales pose significant challenges from both scientific computing \citep{Branets2009} and operator learning \citep{HANO} perspectives. Other notable examples include the \emph{Navier-Stokes equation}, which models fluid flow and exhibits turbulence behavior at high Reynolds numbers, and the \emph{Helmholtz equation}, which models time-harmonic acoustic waves and is challenging to solve in the high wave number regime. In these multiscale PDEs, the accurate prediction of physical phenomena and properties necessitates the resolution of high-frequency components.

\paragraph{Numerical Methods for Multiscale PDEs}
Multiscale PDEs, even with fixed parameters, present a challenge for classical numerical methods, as their computational cost typically scales inversely proportional to the finest scale $\varepsilon$ of the problem. To overcome this issue, multiscale solvers have been developed by incorporating microscopic information to achieve computational cost independent of $\varepsilon$. One such technique is \emph{numerical homogenization} \citep{Engquist2008}, which identifies low-dimensional approximation spaces adapted to the corresponding multiscale operator. Similarly, fast solvers like \emph{multilevel/multigrid methods} \citep{Hackbusch1985,XuZikatanov:2017} and \emph{wavelet-based multiresolution methods} \citep{Brewster1995,Beylkin1998} may face limitations when applied to multiscale PDEs \citep{Branets2009}, while multilevel methods based on numerical homogenization techniques, such as Gamblets \citep{OwhadiMultigrid:2017}, have emerged as a way to discover scalable multilevel algorithms and operator-adapted wavelets for multiscale PDEs. In recent years, there has been increasing exploration of neural network methods for solving multiscale PDEs despite the spectral bias or frequency principle \citep{rahaman2018spectral, ronen2019the, xu2020frequency} indicating that deep neural networks (DNNs) often struggle to effectively capture high-frequency components of functions. To address this limitation, specifically designed neural solvers \citep{Li_2020, WANG2021113938, li2021subspace} have been developed to mitigate the spectral bias and accurately solve multiscale PDEs (with fixed parameters).


\subsection{Neural Operator for Multiscale PDEs}
\label{sec:background:neuralsolver}

Neural operators, as proposed by \citet{li2020fourier, gupta2021multiwavelet}, have shown great promise in capturing the input-output relationship of parametric partial differential equations (PDEs). However, multiscale PDEs introduce new challenges for neural operators. Fourier or wavelet transforms, which are central to the construction of \citet{li2020fourier, gupta2021multiwavelet}, may not always be effective, even for multiscale PDEs with fixed parameters. Moreover, while universal approximation theorems exist for FNO-type models \citep{kovachki2021universal}, achieving a meaningful convergence rate often requires "excessive smoothness" that may be absent for multiscale PDEs. Additionally, aliasing errors becomes significant in multiscale PDEs \citep{bartolucci2023neural}, raising concerns about continuous-discrete equivalence. The work by \cite{HANO} addresses the issue of spectral bias in (multiscale) operator learning and highlights the challenges faced by existing neural operators in capturing high-frequency components of multiscale PDEs. These neural operators tend to prioritize the fitting of low-frequency components over high-frequency ones, limiting their ability to accurately capture fine details. To overcome this limitation, \cite{HANO} proposes a new architecture for multiscale operator learning that leverages hierarchical attention mechanisms and a tailored loss function. While these innovations help reduce the spectral bias and improve the prediction of multiscale solutions, it is worth noting that hierarchical attention induces a significant computational cost.

\subsection{Dilated convolution}

In this paper, we focus on utilizing dilated convolutions to capture the high-resolution components of the data due to their simplicity and efficiency. Dilated convolution, also known as atrous convolution, was initially developed in the ``algorithme \`{a} trous" for wavelet decomposition \citep{Holscheider_book_dilation}. Its primary purpose was to increase image resolution and enable dense feature extraction without additional computational cost in deep convolutional neural networks (CNNs) by inserting "holes" or zeros between pixels in convolutional kernels. By incorporating dilated convolution, networks can enlarge receptive fields, capture more global information, and gather contextual details, which are crucial for dense prediction tasks such as semantic segmentation. Various approaches have been proposed to leverage dilated convolution for this purpose \citep{Yu2015MultiScaleCA,wang2018HDC}, and have demonstrated comparable results compared with U-Net and attention based models.

More recently, dilated convolution has also found applications in operator learning, such as the Dil-ResNet used for simulating turbulent flow \citep{stachenfeld2022learned}. However, our work demonstrates that using dilated convolution alone is not sufficient to accurately capture the solution. Instead, we propose an interwoven global-local architecture of Fourier layers with dilated convolution layers. Furthermore, while Dil-ResNet requires up to 10 million training steps to achieve satisfactory results, our model offers a more efficient approach. It is worth noting that there are many alternative approaches to extract multiscale features, inspired by developments in numerical analysis and computer vision. These include hierarchical matrix methods \citep{fan2019multiscale}, hierarchical attention \citep{liu2021swin,HANO}, U-Net \citep{ronneberger2015u} and U-NO \citep{ashiqur2022u}, wavelet-based methods \citep{gupta2021multiwavelet}, among others. In  Section \ref{sec:experiments}, we will conduct a comprehensive benchmark of these different multiscale feature extraction techniques to evaluate their performance.

\section{Methods}

We adopt a data-driven approach to approximate the operator $\mathcal{S}: \mathcal{H}_1 \mapsto \mathcal{H}_2$ as in references \cite{li2020fourier, cao2021choose, HANO}. The operator $\mathcal{S}$ maps between two infinite-dimensional Banach spaces $\mathcal{H}_1$ and $\mathcal{H}_2$, and aims to find the solution to the parametric partial differential equation (PDE) $\mathcal{L}_\va(\vu) = f$, where the input/parameter $\va\in\mathcal{H}_1$ is drawn from a distribution $\mu$, and the corresponding output/solution $\vu\in\mathcal{H}_2$.

To be specific, in this paper, our objective is to address the following operator learning problems:
\begin{itemize}
\item Approximating the nonlinear mapping $\mathcal{S}: \va \mapsto \vu:=\mathcal{S}(\va)$ from the varying parameter $\va$ to the solution $\vu$.
\item Solving the inverse coefficient identification problem, which involves recovering the coefficient from a noisy measurement $\hat{\vu}$ of the solution $\vu$. In this scenario, we aim to approximate $\mathcal{S}^{-1}: \hat{\vu} \mapsto \va:=\mathcal{S}^{-1}(\hat{\vu})$.
\end{itemize}

\begin{figure*}[h]
    \centering
    \begin{minipage}{0.73\textwidth}
        \includegraphics[width=.95\linewidth]{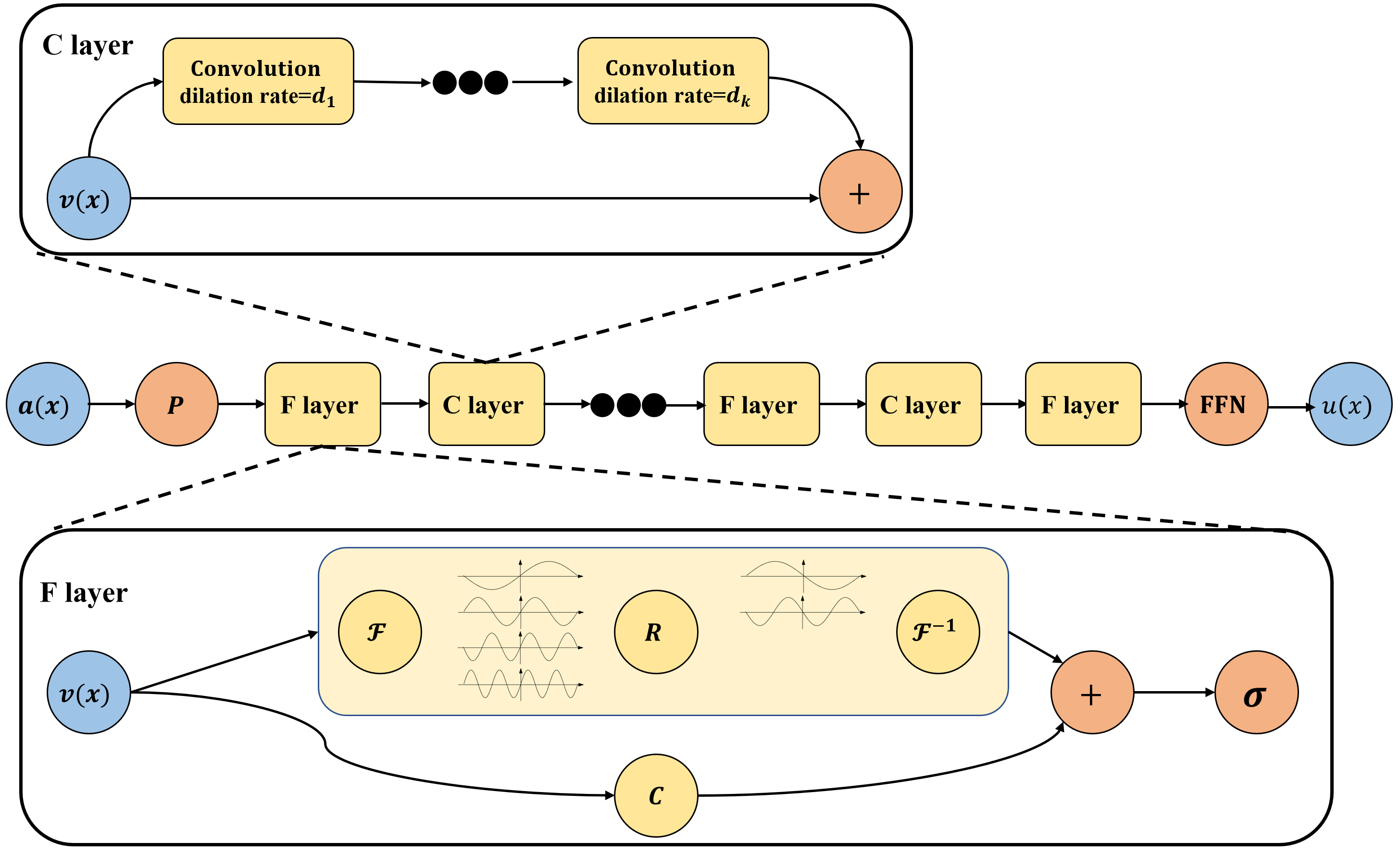}
        \caption{ The architecture of the DCNO neural operator. }
        \label{fig:model} 
    \end{minipage}
    \begin{minipage}{0.24\textwidth}    
    \vspace{-5pt}
    \includegraphics[width=.9\linewidth]{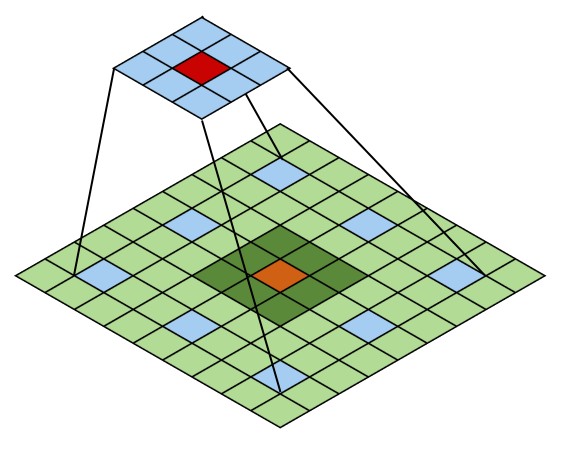}   
    \caption{An example of a two-layer dilated convolution with dilation rates (1, 3).}
        \label{fig:dilation} 
        \vspace{-5mm}
    \end{minipage}
\end{figure*}

\subsection{Model architectures}
In our model, we employ an Encode-Process-Decode architecture \citep{pmlr-v80-sanchez-gonzalez18a, pmlr-v119-sanchez-gonzalez20a}, as shown in Figure \ref{fig:model}.

\begin{itemize}
\item The encoder incorporates a patch embedding function denoted as P, which utilizes a convolutional neural network (CNN). This step is performed to lift the input $a(x)$ to a higher-dimensional channel (feature) space.
\item 
The processor part of the model comprises alternating Fourier layers (F layers) and Convolution layers (C layers). The role of the F layer is to approximate the low-frequency components, while the C layer is responsible for extracting high-frequency features. This alternating approach allows the DCNO model to effectively handle both low-frequency and high-frequency components present in the data. To gain further insights into the influence of the F and C layers, an ablation study is conducted, as described in Section \ref{sec:hyperparameter study(architecture)}. This study provides information about the impact of these layers on the model's performance and enhances our understanding of their significance within the overall architecture.
\item 
The decoder in our model adopts a structure similar to the one proposed in \cite{cao2021choose}. Its primary role is to project high dimensional feature $v(x)$ back to the target dimension of solutions. The decoder comprises two F layers, with the final F layer omitting an activation function, followed by a three-layer feedforward neural network ($\mathit{FFN}$). 
\end{itemize}

\paragraph{F layers:}
The Fourier layers, as proposed in \cite{li2020fourier}, consist of two main components that operate on the input $v(x)$. In the first component, the input undergoes the Fourier transform, followed by a linear transform $\mathit{R}$ acting only on the lower Fourier modes while filtering out the higher modes. The modified input is then obtained by applying the inverse Fourier transform. The first component of the Fourier layer aims to preserve low-frequency global information while reducing the influence of high-frequency components. The second component of the Fourier layers incorporates a local (pointwise) linear transform $\mathit{C}$ to extract high-frequency information. Additionally, the outputs from both components are combined using the GELU activation function.

It is important to note that while the second component of the F layer helps capture some high-frequency details, relying solely on this part is not sufficient to accurately capture high-frequency information. This limitation is why the Fourier neural operator ($\mathit{FNO}$) approach may not perform well for multiscale PDE problems. 

\paragraph{C layers:}
Each convolution layer includes three convolutional neural networks, each utilizing a kernel of size 3 and followed by a GELU activation function. These convolutional neural networks employ dilation rates of $(1, 3, 9, 3, 1)$. The dilation rate determines the spacing between the points with which each point is convolved. A dilation rate of 1 corresponds to a regular convolution where each point is convolved with its immediate neighbors. Larger dilation rates, such as 3 and 9, expand the receptive field of each point to include more distant points. Figure \ref{fig:dilation} illustrates an example of a two-layer dilated convolution with dilation rates of $(1, 3)$. In the first layer, each point is convolved with its neighbors at a distance of 1, and in the second layer, each point is convolved with its neighbors at a distance of 3. As a result, the central red cell has a $9 \times 9$ receptive field. By incorporating multiple dilation rates, the model can capture long-range dependencies and maintain communication between distant points. This approach enhances the model's ability to capture both local and global information. Residual connections are applied to alleviate the vanishing gradient problem.

In summary, the combination of convolution layers with multiple dilation rates effectively enlarges the receptive fields of the network, facilitating the aggregation of global information and leveraging the advantages of convolutions in extracting localized features (see Section \ref{sec:Inverse problem}). However, in most cases in operator learning, relying solely on convolutional layers may not yield satisfactory results, and the combination with Fourier layers can boost performance. See the ablation results in Section \ref{sec:hyperparameter study(architecture)} for more details.

\section{Experiments}
\label{sec:experiments}


In this section, we present a series of numerical experiments comparing DCNO with different operator models based on several metrics, including relative $L^2$ error, parameter count, memory consumption, and training time per epoch. The goal of these experiments is to assess the performance of various operator models through different types of tasks. Specifically, we investigate the performance of these models in the context of multiscale elliptic equations, time-dependent Navier-Stokes equations, inverse coefficient identification for multiscale elliptic equations and the Helmholtz equation. The results consistently demonstrate that the DCNO model outperforms other operator models. It achieves superior accuracy, robustness, and cost-accuarcy trade-off in all the considered scenarios.

\subsection{Data generation}
\label{sec:data generation}
Multiscale elliptic equations are a fundamental class of problems, exemplified by the following second-order elliptic equation in divergence form:
$$
\left\{
    \begin{aligned}
        -\nabla \cdot (a(x) \nabla u(x)) &= f(x), & & x \in D \\
        u(x) &= 0, & & x \in \partial D
    \end{aligned}
\right.
\label{eqn:darcy} 
$$
Here, the coefficient $a(x)$ satisfies $0 < a_{\min} \leq a(x) \leq a_{\max}$ for all $x \in D$, and $f \in H^{-1}(D;\mathbb{R})$ represents the forcing term. The coefficient-to-solution map is denoted as $\mathcal{S}: L^{\infty}(D;\mathbb{R}^{+}) \rightarrow H_{0}^{1}(D;\mathbb{R})$, where $u = \mathcal{S}(a)$. The coefficient $a(x)$ may exhibit rapid oscillations (e.g., $a(x) = a(x/\varepsilon)$ with $\varepsilon \ll 1$), high contrast ratios with $a_{\max}/a_{\min} \gg 1$, and even a continuum of non-separable scales. Handling rough coefficients poses significant challenges from both scientific computing \citep{Branets2009} and operator learning perspectives. In following, we give the details of the two examples of multiscale elliptic equations benchmarked in this paper.

\subsubsection{Darcy rough example}
\label{sec:darcy}

Darcy flow, originally introduced by Darcy \citep{darcy1856fontaines}, has diverse applications in modeling subsurface flow pressure, linearly elastic material deformation, and electric potential in conductive materials. The two-phase coefficients and solutions are generated using the approach outlined in \url{https://github.com/zongyi-li/fourier_neural_operator/tree/master/data_generation}. 
Given the computational domain $[0,1]^2$, the coefficients $a(x)$ are generated according to 
$$a\sim\mu:=\psi_{\#} \mathcal{N}\left(0,(-\Delta+ c I)^{-2}\right),$$ 
where $\Delta$ represents the Laplacian with zero Neumann boundary condition. The mapping $\psi: \mathbb{R} \rightarrow \mathbb{R}$ takes the value $12$ for the positive part of the real line and $2$ for the negative part, with a contrast of $6$. The push-forward is defined in a pointwise manner. These generated datasets serve as benchmark examples for operator learning in various studies, including \citet{li2020fourier}, \citet{gupta2021multiwavelet}, and \citet{cao2021choose}. The parameter $c$ can be used to control the "roughness" of the coefficient and corresponding solution. In the aforementioned references, the parameter $c$ is set as $c=9$, while in \citet{HANO}, a value of $c=20$ is used to generate a rougher coefficient. The forcing term is fixed as $f(x)\equiv 1$. Solutions $u$ are obtained using a second-order finite difference scheme on a $512 \times 512$ grid. Lower-resolution datasets are created by sub-sampling from the original dataset through linear interpolation.

\subsubsection{Multiscale trigonometric example}
\label{sec:mul}
Multiscale trigonometric coefficients are described in \citet{OwhadiMultigrid:2017}, as an example of highly oscillatory coefficients. Given the domain $D$ $[-1,1]^2$, the coefficient $a(x)$ is specified as follows:
$$a(x) = \prod \limits_{k=1}^6 \left(1+\frac{1}{2} \cos(a_k \pi (x_1+x_2))\right)\left(1+\frac{1}{2} \sin(a_k \pi (x_2-3x_1))\right)$$
Here, $a_k$ is uniformly distributed in $[2^{k-1},1.5\times 2^{k-1}]$. The forcing term is fixed at $f(x)\equiv 1$.
To obtain the reference solutions, the $\mathcal{P}_1$ Finite Element Method (FEM) is employed on a $1023 \times 1023$ grid. Lower-resolution datasets are generated by downsampling the higher-resolution dataset through linear interpolation.

\begin{figure}[H]
    \centering
    \subfigure[coefficient $a(x)$]{\includegraphics[width=0.3\textwidth]{MC_mul_coeff2.png}}
    \subfigure[1D slices of the predicted solutions]{\includegraphics[width=0.3\textwidth]{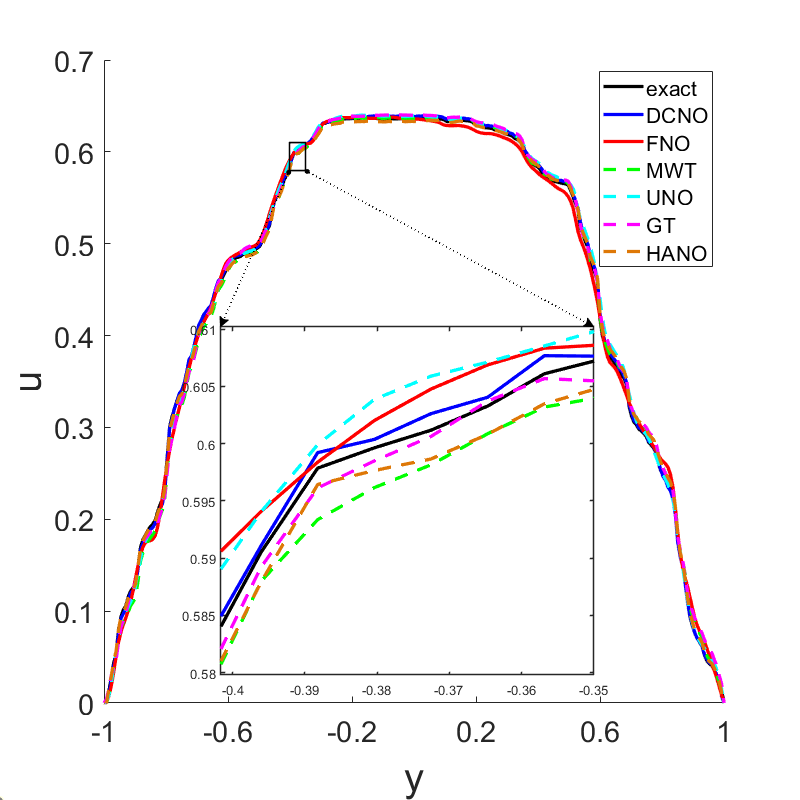}}
    \subfigure[1D slices of the predicted solution derivatives]{\includegraphics[width=0.3\textwidth]{gamblet_slides_grad2.png}}
    \caption{(a) multiscale trigonometric coefficient, (b)comparison of predicted solutions on the slice $x=0$, (c) comparison of predicted derivative $\frac{\partial u}{\partial y}$ on the slice $x=0$.}
    \label{fig:gamblet slices} 
\end{figure}

We present the multiscale trigonometric coefficient, reference solution, and a comparison with other operator learning models in Figure \ref{fig:gamblet slices}. Among the models considered, DCNO demonstrates superior accuracy in predicting function values and, more importantly, accurately captures the fine-scale oscillations. This is evident in the predicted derivatives shown in (c) of Figure \ref{fig:gamblet slices}.

\subsection{Training and evaluation setup}
\label{sec:setup}
Unless stated otherwise, the train-val-test split datasets used consist of 1000, 100, and 100 samples, respectively, with a maximum of 500 training epochs and batch size of 8. The Adam optimizer is utilized with a decay of $\mathrm{1e}-4$ and a 1cycle learning rate scheduler \citep{LNSmith20191cycle}. For the Navier-Stokes equations, the dataset resolution is set to $64\times 64$, again with a maximum of 500 epochs and batch size of 32. All experiments are executed on an NVIDIA A100 GPU.

\subsection{Multiscale Elliptic Equations}
\label{sec:experiments:multiscale elliptic}

We examine the effectiveness of the DCNO model on the multiscale elliptic equation, given by the following  second-order linear elliptic equation,
\begin{equation}
    \left\{
	\begin{aligned}
		-\nabla \cdot(a (x) \nabla u (x)) &=f(x), & & x \in D \\
		u (x) &=0, & & x \in \partial D
	\end{aligned}
    \right.
\label{eqn:muL^2d}
\end{equation}
with rough coefficients and Dirichlet boundary conditions. Our goal is to approximate the operator $\mathcal{S}:L^\infty (D;\mathbb{R}_+) \rightarrow H_0^1(D;\mathbb{R})$, which maps the coefficient function $a(x)$ to the corresponding solution $u$. We assess the model on two-phase Darcy rough coefficients (Darcy rough) given in \citet{HANO} where the coefficients are significantly rougher compared to the well-known benchmark proposed in \citet{li2020fourier}. We also consider multiscale trigonometric coefficients with higher contrast, following the setup in \citet{OwhadiMultigrid:2017,HANO}. The coefficients and solutions are displayed in Figure \ref{fig:muL^2d}.

\begin{figure*}[h]
    \centering
    \begin{minipage}{0.7\textwidth}
        \subfigure[\scriptsize coefficient]{
        \begin{minipage}[b]{0.2\textwidth}
            \includegraphics[width=\textwidth,height=0.85\textwidth]{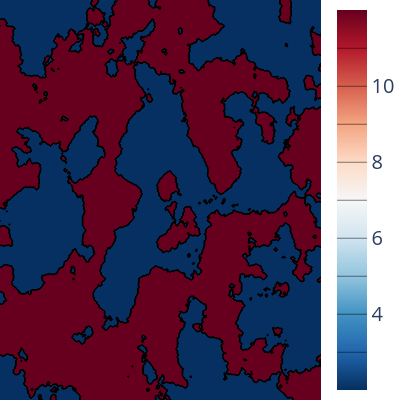}
            \includegraphics[width=1.05\textwidth,height=0.85\textwidth]
            {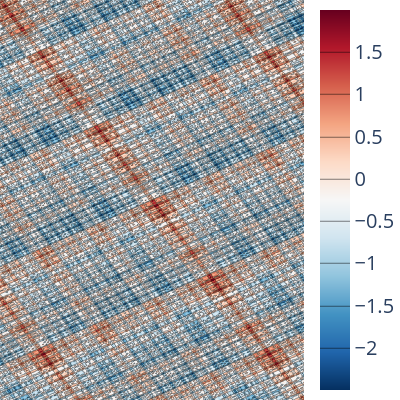}
        \end{minipage}
        }
        \subfigure[\scriptsize reference solution]{
            \begin{minipage}[b]{0.2\textwidth}
                \includegraphics[width=\textwidth,height=0.85\textwidth]{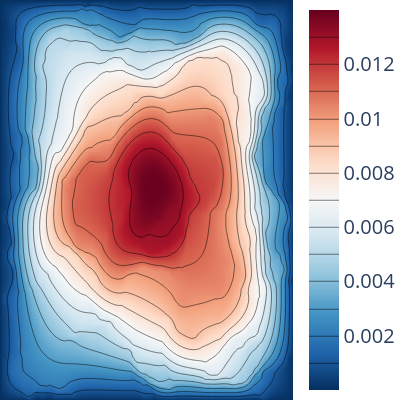}
                \includegraphics[width=0.935\textwidth,height=0.85\textwidth]{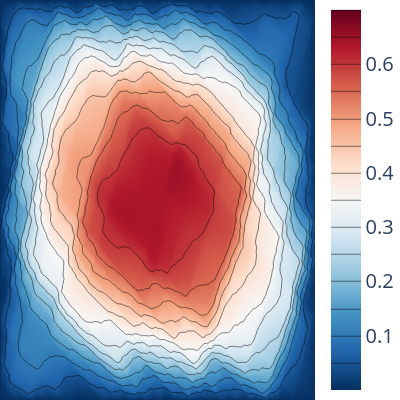}
            \end{minipage}
        }
        \subfigure[\scriptsize abs. error of DCNO (in $\log_{10}$ scale)]{
            \begin{minipage}[b]{0.2\textwidth}
                \includegraphics[width=\textwidth,height=0.85\textwidth]{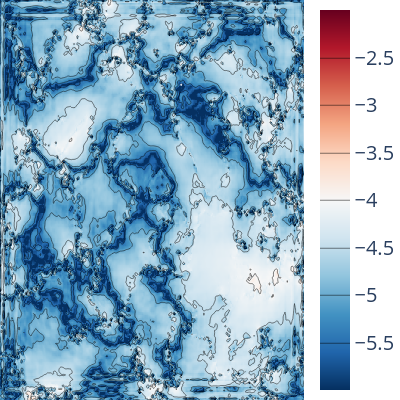}
                \includegraphics[width=1.01\textwidth,height=0.85\textwidth]{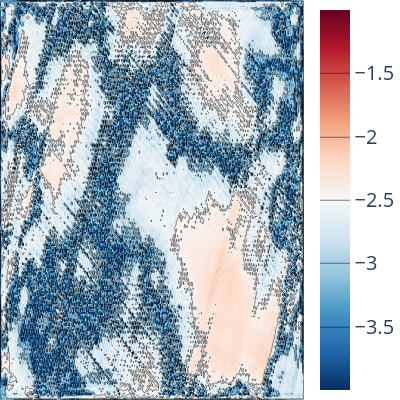}
            \end{minipage}
        }
        \subfigure[\scriptsize abs. error of FNO (in $\log_{10}$ scale)]{
            \begin{minipage}[b]{0.2\textwidth}
                \includegraphics[width=\textwidth,height=0.85\textwidth]{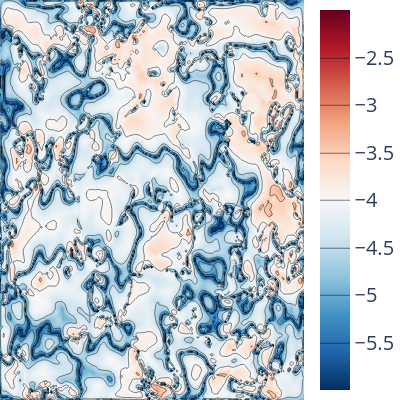}
                \includegraphics[width=1.01\textwidth,height=0.85\textwidth]{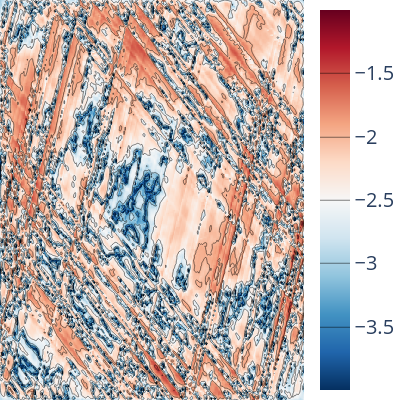}
            \end{minipage}
        }
    \label{fig:muL^2d}
    \caption{ \textbf{Top:} Darcy rough example, (a) coefficient, (b) reference solution,  (c) DCNO, absolute (abs.) error, (d) FNO, abs. error;  \textbf{Bottom:} multiscale trignometric example, (a) coefficient (in $\log_{10}$ scale),  (b) reference solution, (c) DCNO, abs. error (in $\log_{10}$ scale), (d) FNO, abs. error (in $\log_{10}$ scale). }
    \end{minipage}
    \quad
    \begin{minipage}{0.25\textwidth}    
    \begin{minipage}[b]{0.9\textwidth}
            \includegraphics[width=1.1\textwidth,height=0.82\textwidth]{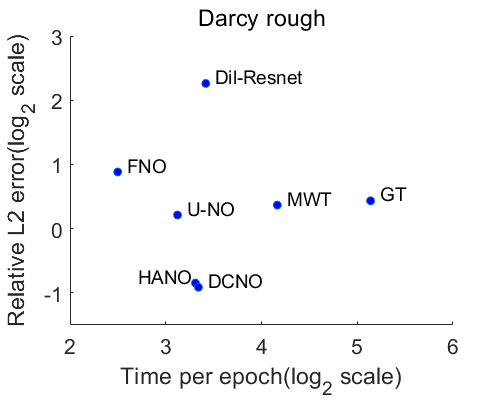}
            \includegraphics[width=1.1\textwidth,height=0.82\textwidth]{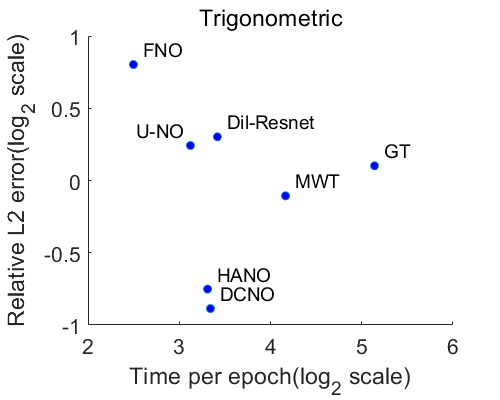}
        \end{minipage}
    \caption{Cost-accuracy (in $\log_{2}$ scale) trade-off}
    \label{fig:tradeoff}
    \end{minipage}
\end{figure*}

The experimental results for multiscale elliptic equations conducted at different resolutions are presented in Table \ref{tab:multi elliptic}, and can be summarized below,
\begin{itemize}
    \item DCNO achieves the lowest relative error compared to other neural operators at various resolutions, and the errors remain approximately invariant with the resolution. Compared to FNO, DCNO has a remarkable accuracy improvement of 71$\%$ and 69$\%$ in the cases of Darcy rough and multiscale trigonometric. Furthermore, we observe that DCNO achieves the best cost-accuracy trade-off among all the neural operators we tested in Figure \ref{fig:tradeoff}, where the cost is measured by the training time per epoch for $s=256$ (input resolution) of the Darcy rough example. 
    \item Dil-ResNet, with its convolutional architecture, has the fewest parameters among all models. However, it requires more memory and longer training time per epoch compared to DCNO and FNO. Furthermore, its accuracy is not as ideal as the other models.
\end{itemize}
These results highlight the superior performance of the DCNO model in terms of accuracy and efficiency, making it a favorable choice for solving multiscale elliptic equations.

\begin{table}[H]
    \centering
    \scalebox{0.8}{
        \begin{tabular}{l|c|c|c|cc} 
        \toprule
        
         & \textbf{Parameters} & \textbf{Memory} & \textbf{Time per}&{\textbf{Darcy rough}} & {\textbf{Trigonometric}}\\ 
         \textbf{Model} & $\times 10^{6}$     & (GB)              & \textbf{epoch}(s)
        & &\\               
        \toprule
         \textsc{FNO}  & $2.37$ & $1.79$ & \bm{$5.65$} & $1.749 \pm 0.029$ & $1.744 \pm 0.034$ \\
        \textsc{MWT}  & $9.81$ & $2.54$ & $17.94$ & $1.289 \pm 0.043$ & $0.929 \pm 0.029$ \\
        \textsc{U-NO}  & $16.39$ & \bm{$1.45$} & $8.71$ & $1.159 \pm 0.021$ & $1.183 \pm 0.018$ \\
        \textsc{GT}  & $2.22$ & $9.32$ & $35.25$ & $1.351 \pm 0.030$ & $1.073 \pm 0.013$\\
        \textsc{HANO}  & $8.65$ & $2.33$& $9.92$ & $0.556 \pm 0.031$ & $0.594 \pm 0.148$ \\
        \textsc{Dil-ResNet}  & \bm{$0.58$} & $5.71$ & $10.69$ & $4.794 \pm0.175$ & $1.233 \pm 0.050$\\
        \hline
        \textsc{DCNO}  & $2.51$ & $3.94$ & $10.14$ & $\bm{0.531} \pm 0.026$ & $\bm{0.541} \pm 0.051$\\
        
        \toprule
        \end{tabular}
    }
    \caption{Benchmarks on multiscale elliptic equations at various input resolution $s$. Performance are measured with relative $L^2$ errors ($\times 10^{-2}$), number of parameters, memory consumption for a batch size of 8, and time per epoch for $s=256$ of the Darcy rough example during the training process.}
    
    \label{tab:multi elliptic}
\end{table}

\subsubsection{Hyperparameter Study(dilated convolution)}
\label{sec:hyperparameter study(dilated convolution)}

\begin{table}[H]
    \centering
    \scalebox{0.8}{
    \begin{tabular}{l|ccc}
    \hline
    \textbf{Dilation} &\textbf{Parameters for C layers} & \textbf{Darcy} $(\times 10^{-2})$ & \textbf{Trigonometric} $(\times 10^{-2})$\\
    \hline
    $(1, 3, 9, 3, 1)$ & $138720$ & $0.531$  & $0.541$\\
    $(1, 1, 1, 1, 1)$ & $138720$ & $0.792$  & $0.650$\\
    $(1, 3, 9)$ & $83232$ & $0.623$ & $0.814$\\
    $(1, 3, 1)$ & $83232$ & $0.691$  & $0.641$\\
    $(1, 1, 1)$ & $83232$ & $0.768$  & $0.805$\\
    $(1)$  &$27744$ & $1.065$ & $0.976$  \\
    \hline
    $(1)$ (\textrm{kernel size $9\times 9$})& $248928$ & $0.778$ & $0.627$\\
    \hline
    \end{tabular}
    }
    \caption{Hyperparameter study: impact of dilation rate and kernel size}
    \label{tab:Hyperparameter study}
\end{table}
We conducted a hyperparameter study to investigate the influence of different dilation rates and kernel sizes in the C layers of the DCNO model for multiscale elliptic PDEs. Those parameters will in turn determine the reception field and parameter count of the convolution. The results of this study are summarized in Table \ref{tab:Hyperparameter study}. The default size of the convolutional kernel is $3 \times 3$, unless otherwise specified. The dilation rates determine the configuration of the C layers in the DCNO model, where dilation rates of $(1, 3, 9)$ means that the C layers consist of three dilated convolutions with dilation factors of 1, 3, and 9.
As expected, increasing the number of layers in the C layers leads to improved results. This can be attributed to the fact that additional layers can capture more complex features, thereby enhancing the model's accuracy. To assess the impact of hierarchical dilated convolution, we compare the outcomes obtained with dilation rates $(1, 1, 1, 1, 1)$ and $(1, 1, 1)$ against those acquired with dilation rates $(1, 3,9,3,1)$ and $(1, 3, 9)$.
The results clearly demonstrate that hierarchical dilated convolution has a positive effect on the outcomes. This suggests that the ability to capture multiscale information through multiple dilation rates proves beneficial in enhancing the performance of the model.
Additionally, we compared the performance of a single layer with a large $9\times9$ convolutional kernel against a layered dilated convolution with a $3\times3$ kernel size. Interestingly, we found that the latter yielded better results with fewer parameters. This highlights the effectiveness of layered dilated convolutions in capturing complex features while maintaining model efficiency.

\subsubsection{Hyperparameter Study(architecture)}
\label{sec:hyperparameter study(architecture)}
\begin{table}[H]
    \centering
    \scalebox{0.8}{
    \begin{tabular}{l|cc}
    \hline
    \textbf{Model} & \textbf{Darcy rough}  & \textbf{Trigonometric} \\
    \hline
    DCNO(\textrm{FCFCFCF}) & $0.531$ & $0.541$\\
    \hline
    DCNO$^{1}$ (\textrm{FFFFFFF}) & $1.344$ & $1.256$\\
    DCNO$^{2}$ (\textrm{CCCCCCC}) & $4.324$ & $1.045$\\
    \hline
    DCNO$^{3}$ (\textrm{CCCCCCF}) & $0.594$ & $0.842$\\
    DCNO$^{4}$ (\textrm{FCCCCCC}) & $1.842$ & $0.923$\\
    DCNO$^{5}$ (\textrm{CFFFFFF}) & $0.707$ & $0.732$\\
    DCNO$^{7}$ (\textrm{FFFFFFC}) & $1.180$ & $0.864$\\
    \hline
    DCNO$^{8}$ (\textrm{FFFFCCC}) & $1.240$ & $0.631$\\
    DCNO$^{9}$ (\textrm{CCCFFFF}) & $0.676$ & $0.872$\\
    \hline
    \end{tabular}
    }
    \caption{Ablation Study of DCNO: different combinations of F- and C- layers. }
    \label{tab:ablation1}
\end{table}

\begin{figure}[h]
    \centering
    \subfigure[\scriptsize absolute error spectrum of DCNO in $\log_{10}$ scale]{
    \begin{minipage}[b]{0.16\textwidth}
        \includegraphics[width=\textwidth,height=0.85\textwidth]{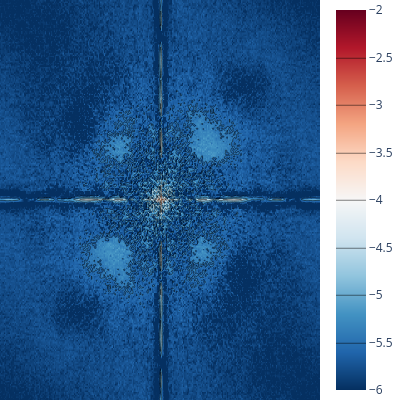}
        \includegraphics[width=\textwidth,height=0.85\textwidth]
        {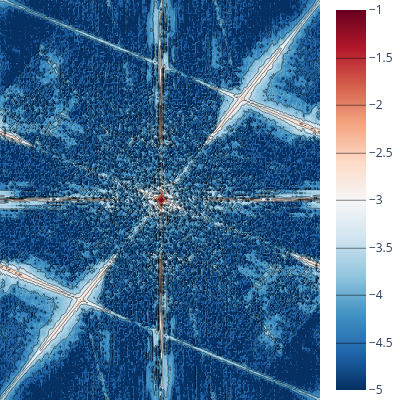}
    \end{minipage}
    }
    \subfigure[\scriptsize absolute error spectrum of DCNO$^{1}$ (only F layers) in $\log_{10}$ scale]{
        \begin{minipage}[b]{0.16\textwidth}
           \includegraphics[width=\textwidth,height=0.85\textwidth]{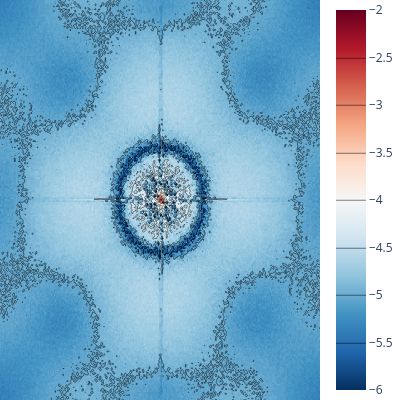}
            \includegraphics[width=\textwidth,height=0.85\textwidth]
            {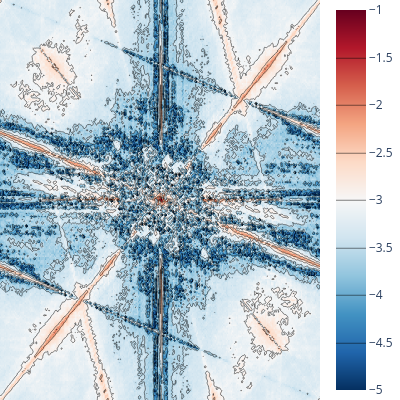}
        \end{minipage}
    }
    \subfigure[\scriptsize absolute error spectrum of DCNO$^{2}$ (only C layers) in $\log_{10}$ scale]{
        \begin{minipage}[b]{0.16\textwidth}
            \includegraphics[width=\textwidth,height=0.85\textwidth]{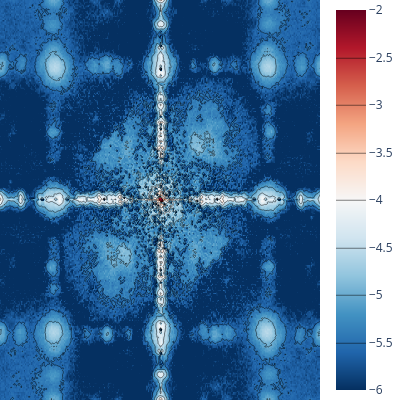}
            \includegraphics[width=\textwidth,height=0.85\textwidth]
            {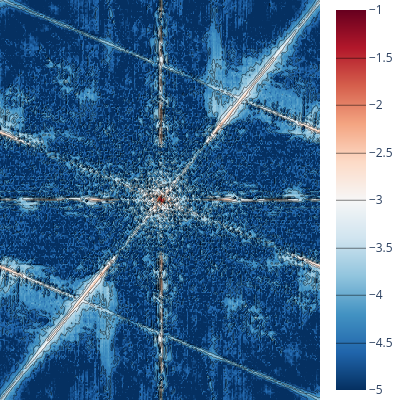}
        \end{minipage}
    }
    \subfigure[\scriptsize (testing) low frequency error dynamics in spectral domain]{
        \begin{minipage}[b]{0.16\textwidth}
            \includegraphics[width=\textwidth,height=0.85\textwidth]{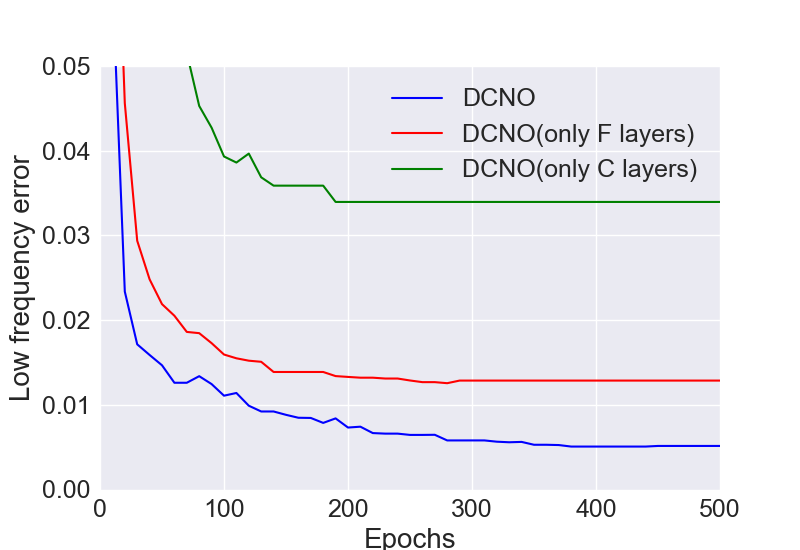}
            \includegraphics[width=\textwidth,height=0.85\textwidth]{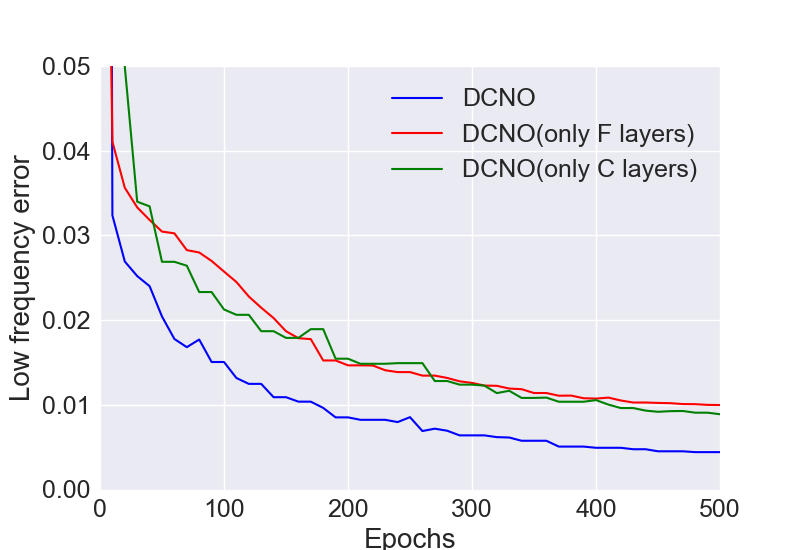}
        \end{minipage}
    }
    \subfigure[\scriptsize (testing) high frequency error dynamics in spectral domain]{
        \begin{minipage}[b]{0.16\textwidth}
            \includegraphics[width=\textwidth,height=0.85\textwidth]{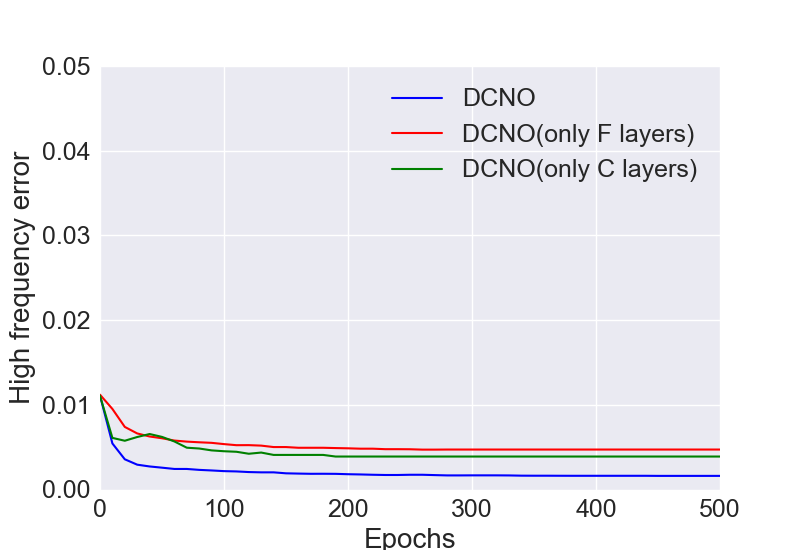}
            \includegraphics[width=\textwidth,height=0.85\textwidth]{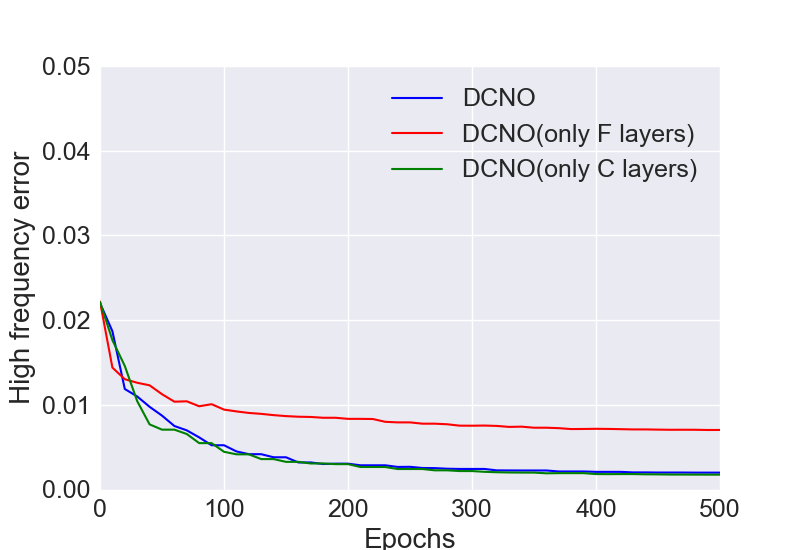}
        \end{minipage}
    }
    \label{fig:muL^2d}
    \caption{ \textbf{Top:} Darcy rough example, \textbf{Bottom:} multiscale trignometric example.}
\end{figure}

To assess the impact of different components of the DCNO model on multiscale elliptic PDEs, we conducted experiments testing various combinations of F- and C-layers.

DCNO$^{1}$ is a variant of the DCNO model where all C layers are replaced with F layers, akin to an enhanced version of the FNO model. However, we observed that DCNO$^{1}$ does not achieve comparable accuracy to the full DCNO model. In contrast, DCNO$^{2}$ replaces all F layers with C layers within the DCNO framework. Interestingly, we found that DCNO$^{2}$ demonstrates significantly better performance in the multiscale trigonometric case compared to the Darcy rough case. This observation suggests that dilated convolutions in the C layers effectively enhance results when dealing with rough coefficients. The C layers' capability to capture local interactions and finer details appears particularly advantageous in scenarios with pronounced coefficient variations.

While DCNO$^{3}$, DCNO$^{4}$, DCNO$^{5}$, and DCNO$^{6}$ each incorporate only one F layer or C layer compared to DCNO$^{1}$ and DCNO$^{2}$, their mixed utilization of both F and C layers enables them to outperform DCNO$^{1}$ with only F layers and DCNO$^{2}$ with only C layers. This underscores the critical importance of combining F and C layers to achieve superior performance. F layers are adept at capturing global information and long-range dependencies, while C layers excel in capturing local interactions and finer details. Both components are essential for accurately modeling the complex dynamics of the system.

To underscore the importance of interleaved architecture, we scrutinized DCNO$^{7}$ and DCNO$^{8}$. DCNO$^{7}$ features three C layers following four F layers, while DCNO$^{8}$ places three C layers before four F layers. However, both DCNO$^{7}$ and DCNO$^{8}$ proved less effective than the complete DCNO model. This suggests that the specific interleaving of C and F layers within the DCNO model's architecture is critical for achieving optimal performance.

In summary, the comparison among the DCNO variants underscores the critical roles played by both F layers and C layers within the DCNO model. These components are essential and complement each other in achieving the superior performance demonstrated by the complete DCNO model. F layers excel in capturing global information and long-range dependencies, whereas C layers are adept at capturing local interactions and finer details. The interleaved architecture of the DCNO model, with a carefully ordered combination of F and C layers, is pivotal for achieving its outstanding performance.

\subsubsection{Spectral bias}
\label{sec:spectralbias}
\begin{figure}[H]
    \centering
    \subfigure[DCNO]{\includegraphics[width=0.2\textwidth]{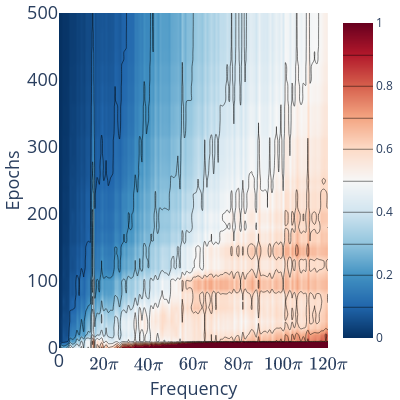}}
    \subfigure[FNO]{\includegraphics[width=0.2\textwidth]{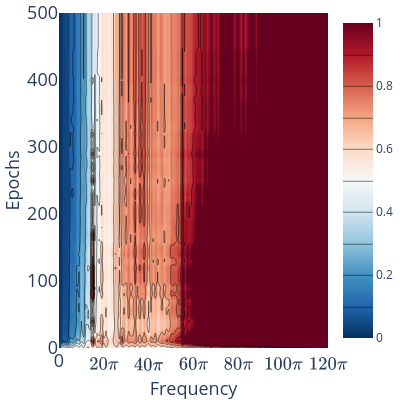}}
    \subfigure[MWT]{\includegraphics[width=0.2\textwidth]{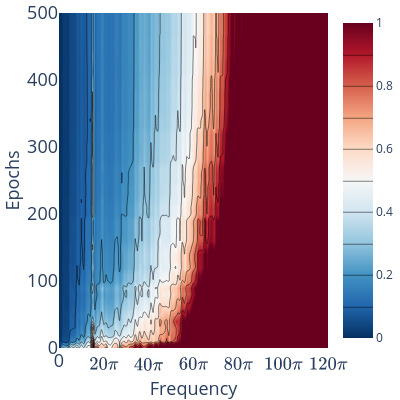}}
    \subfigure[U-NO]{\includegraphics[width=0.2\textwidth]{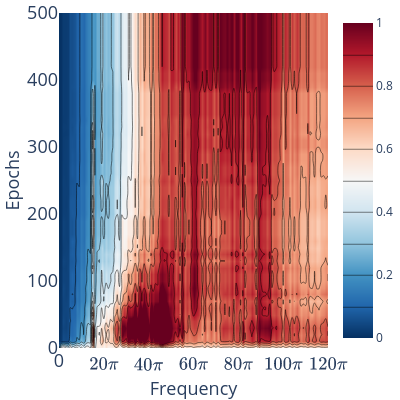}}
    \subfigure[GT]{\includegraphics[width=0.2\textwidth]{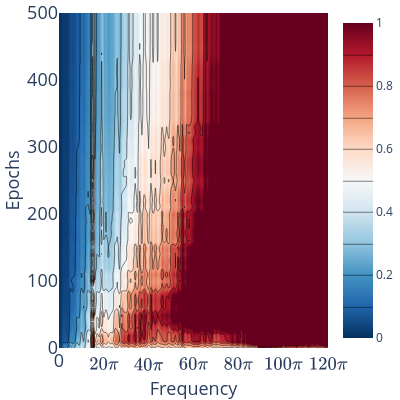}}
    \subfigure[HANO]{\includegraphics[width=0.2\textwidth]{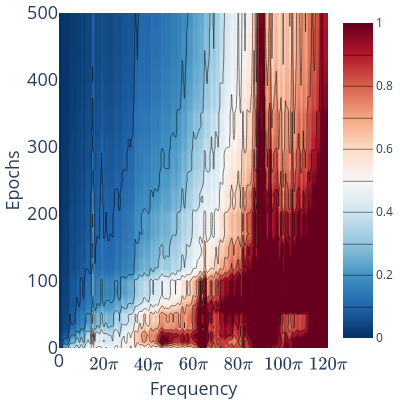}}
    \subfigure[Dil-ResNet]{\includegraphics[width=0.2\textwidth]{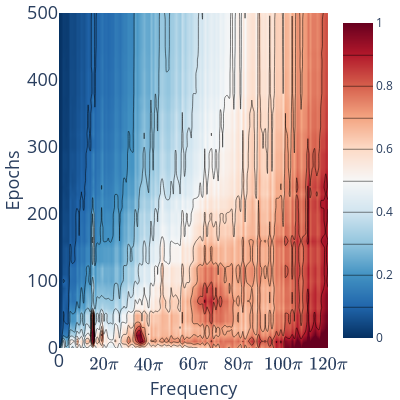}}
    \caption{Error dynamics in the frequency domain for multiscale trigonometric example.}
    \label{fig:special bias}
\end{figure}

The spectral bias, also known as the frequency principle, suggests that deep neural networks (DNNs) face challenges in effectively learning high-frequency components of functions that exhibit variations at multiple scales. This phenomenon has been extensively studied and discussed in the literature \citep{rahaman2018spectral,ronen2019the,xu2020frequency} in the context of function approximation.

In the context of operator learning, we conducted an analysis on the relative error spectrum dynamics of a multiscale trigonometric example presented in Section \ref{sec:experiments:multiscale elliptic}. The analysis is illustrated in Figure \ref{fig:special bias}.
To begin, we computed the Fourier transform of the relative error in the frequency domain $[-128\pi, 128\pi]^2$. Our examination focused on the error density $\rho(r)$ within the annulus $A(r)$, which satisfies the equation $\int_{A(r)} \rho(r)r dr = \sum_{i \in A(r)} \epsilon_{i}$.
The annulus $A(r) := B(r+1) \backslash B(r)$, where $B(r)$ represents a sphere of radius $r$. The term $\epsilon_{i}$ represents the Fourier-transformed relative error at a lattice point indexed by $i$ in the discrete frequency domain.

In Figure \ref{fig:special bias}, the x-axis represents dominant frequencies arranged from low to high frequency, while the y-axis represents the number of training epochs. The plot reveals several significant findings. Firstly, the DCNO model exhibits a faster error decay for higher frequencies, demonstrating its effectiveness in capturing high-frequency components. Additionally, the DCNO model shows a consistent reduction in errors across all frequencies, indicating its capability to learn variations at multiple scales proficiently. Moreover, the DCNO model outperforms other methods by achieving lower testing errors, highlighting its effectiveness in addressing spectral bias and accurately predicting functions with diverse scale variations. This analysis provides compelling evidence of the DCNO model's advantages in operator learning and its ability to handle spectral bias while accurately modeling functions across various frequency ranges.

\subsection{Navier-Stokes Equation}
\label{sec:experiments:NS}

In this section, we focus on the following 2D Navier-Stokes equation in vorticity form on the unit torus $\mathsf{T}$, as benchmarked in \citep{li2020fourier}. 
\begin{equation}
\left\{
\begin{aligned}
\partial_{t} w(x, t)+u(x, t) \cdot \nabla w(x, t) &=\nu \Delta w(x, t)+f(x), & x \in (0,1)^2, t \in (0,T] \\ 
\nabla \cdot u(x, t) &=0, & x \in (0,1)^2, t \in [0,T] \\
w(x, 0) &=w_{0}(x), & x \in (0,1)^2
\end{aligned}
\right.
\end{equation}

In the context of fluid dynamics, the variables used in the equations have the following interpretations:
\begin{itemize}
    \item The velocity field is represented by the symbol $u$.
    \item The vorticity field is denoted as $w$, and it is defined as the curl of the velocity field, i.e., $w=\nabla \times u$.
    \item The initial vorticity distribution is denoted by $w_0$.
    \item The viscosity coefficient is represented by $\nu$.
    \item The forcing term is given by $f(x) = 0.1\left(\sin \left(2 \pi(x_1+x_2)\right) + \cos \left(2 \pi(x_1+x_2)\right)\right)$.
\end{itemize}
The Reynolds number, denoted as $\mathrm{Re}$, is a dimensionless parameter defined as $\mathrm{Re} := \frac{\rho u L}{\nu}$, where $\rho$ is the density (assumed to be $1$), $u$ is the fluid velocity, and $L$ is the characteristic length scale of the fluid (set to $1$). The Reynolds number is inversely proportional to the viscosity coefficient $\nu$. An increase in the Reynolds number tends to promote the transition of the flow to turbulence.
The initial vorticity distribution $w_0(x)$ is generated from a probability measure $\mu$. Specifically, $w_0 \sim \mu$, where $\mu=\mathcal{N}\left(0,7^{3/2}(-\Delta+49 I)^{-2.5}\right)$, and periodic boundary conditions are applied.

The vorticity variable is denoted as $\omega(x,t)$, where $x\in \mathsf{T}$ represents the spatial domain and $t \in [0, T]$ represents the time interval.  Our goal is  to learn the mapping from the vorticity at the previous time step, $w(\cdot, t-1)$, to the vorticity at the current time step, $w(\cdot, t)$. 

During the training process, we follow the approach used in \citet{HANO}. The predicted vorticity at each time step, denoted as $\tilde{w_t}$, is obtained using a recurrence relation: $\tilde{w_t} = \mathcal{G}(\tilde{w}_{t-1})$. Here, $\mathcal{G}$ represents an operator that approximates the mapping from the previous vorticity to the current vorticity. In the training process, we set $\tilde{w}_{i} = w_i$, where $\tilde{w}_{i}$ is the predicted vorticity and $w_{i}$ is the true vorticity at time step $i$. 

During testing, we employ a “rollout" strategy, commonly utilized in fluid dynamics studies \citep{li2020fourier, ashiqur2022u}. This strategy entails predicting the vorticity at each time step using the previously mentioned recurrence relation, with only the initial vorticity $\tilde{w}_{0} = w_0$ known. Our objective is to evaluate the DCNO model and other benchmark operators on their capability to accurately forecast the intricate dynamics of fluid flow over time by assessing their performance in predicting vorticity dynamics.

Table \ref{tab:NS} presents a comprehensive summary of the Navier-Stokes experiment results. For lower Reynolds numbers, FNO, MWT, U-NO, HANO and DCNO achieve similar levels of accuracy. However, as the Reynolds number increases, the DCNO models demonstrate a notable advantage in accuracy. This highlights the effectiveness of the DCNO models in capturing the complex dynamics of fluid flow and accurately predicting vorticity, especially in scenarios with higher Reynolds numbers.
Overall, the results demonstrate that DCNO consistently provides more accurate predictions compared to other methods while maintaining a reasonable computational cost. 

\begin{table}[H]
    \centering
    \scalebox{0.75}{
    \begin{tabular}{l|c|c|c|ccc} 
    \toprule
                   &                     & \textbf{Memory}   & \textbf{Time per}
    & $\nu=1 \mathrm{e}-3$&$\nu=1 \mathrm{e}-4$&$\nu=1 \mathrm{e}-5$\\
    \textbf{Model} & \textbf{Parameters} & \textbf{Requirement}& \textbf{epoch}
    & $T_0=1s$ & $T_0=1s$ & $T_0=1s$ \\    

                   & $\times 10^{6}$     & (GB)              & (s)
    & $T=19s$ & $T=19s$ & $T=19s$ \\               
    \toprule
    
    \textsc{FNO}  & $2.37$ & $0.22$ & \bm{$4.20$} & $4.489 \times 10^{-4}$ & $1.004\times 10^{-2}$ & $6.288 \times 10^{-2}$ \\
    \textsc{FNO}$^{\dag}$  & $25.97$ & $0.99$ & $10.26$ & $6.095 \times 10^{-4}$ & $5.100\times 10^{-3}$ & $3.555 \times 10^{-2}$ \\
    \textsc{MWT}  & $9.81$ & $0.32$ & $27.42$ & $7.075 \times 10^{-4}$ & $1.015 \times 10^{-2}$ & $5.288 \times 10^{-2}$ \\
    \textsc{U-NO}  & $11.91$ & \bm{$0.22$} & $36.97$ & $4.532 \times 10^{-4}$ & $8.192 \times 10^{-3}$ & $6.743 \times 10^{-2}$ \\
    \textsc{HANO}  & $15.37$ & $0.82$ & $18.68$ & $5.734 \times 10^{-4}$ & \bm{$3.759 \times 10^{-3}$} & $2.697 \times 10^{-2}$ \\
    \textsc{Dil-ResNet} & \bm{$0.58$} & $0.71$ & $16.42$ & $2.786\times 10^{-2}$ & $1.259 \times 10^{-1}$ & $2.319 \times 10^{-1}$ \\
    \textsc{DCNO} & $14.40$ & $1.47$ & $16.21$ & \bm{$4.056\times 10^{-4}$} & $3.764 \times 10^{-3}$ & \bm{$2.227 \times 10^{-2}$} \\
    \toprule
    \end{tabular}
    }
    \caption{Benchmarks on Navier Stokes equation.Performance are measured with relative $L^2$ errors , number of parameters, memory consumption for a batch size of 16, and time per epoch for $(\nu=1 \mathrm{e}-5, T_0=10s, T=20s)$ during the training process.}
    \label{tab:NS}
\end{table}

\begin{figure}[H]
        \centering
        \includegraphics[width=15cm]{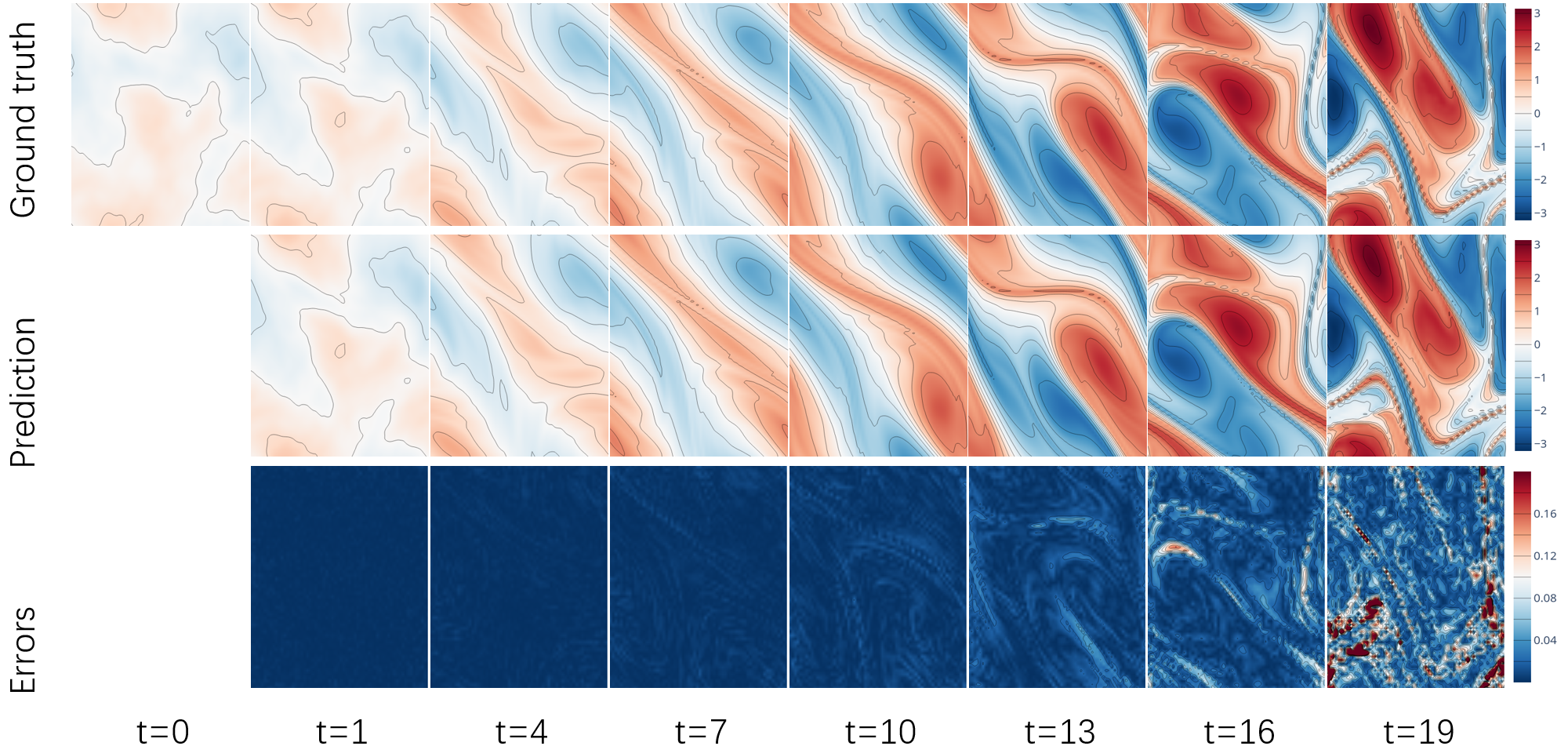}
        \caption{Comparison of Navier Stokes solutions ($\nu=1e-5$) and DCNO predictions.}
        \label{fig:ns_dynamic}
\end{figure}


\subsection{Inverse coefficient identification for multiscale elliptic PDEs}
\label{sec:Inverse problem}

In this section, we address an inverse coefficient identification problem using the same data as the previous example of multiscale elliptic PDE in section \ref{sec:experiments:multiscale elliptic}. Inverse problems play a crucial role in various scientific fields, including geological sciences and medical imaging. In this example, our objective is to learn an approximation to an ill-posed operator $\mathcal{S}^{-1}: H_0^1(D) \mapsto L^\infty(D)$, where $\hat{\vu} = \vu + \epsilon N(\vu) \mapsto \va$. Here, $\epsilon$ represents the level of Gaussian noise added to both the training and evaluation data. The term $N(\vu)$ accounts for the  data-related noise. This task is challenging due to the ill-posed nature of the problem and the presence of noise.

\begin{figure}[H]
    \centering
    \subfigure[]{\includegraphics[width=0.2\textwidth]{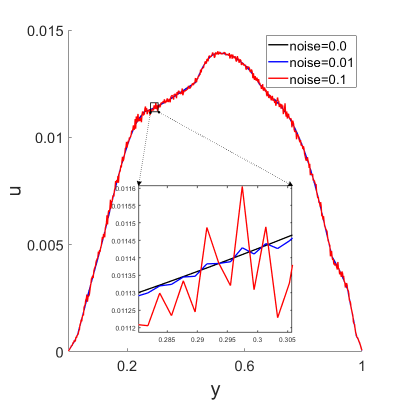}}\hspace{5mm}
    \subfigure[]{\includegraphics[width=0.2\textwidth]{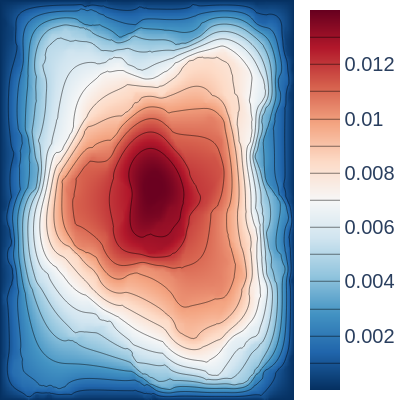}}\hspace{5mm}
    \subfigure[]{\includegraphics[width=0.2\textwidth]{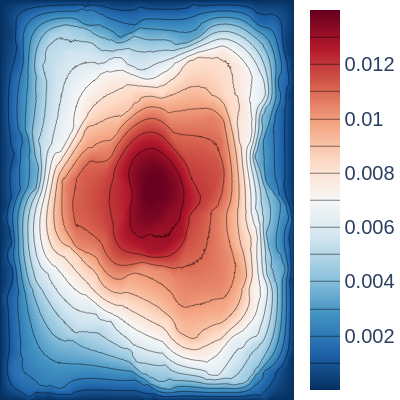}}\hspace{5mm}
    \subfigure[]{\includegraphics[width=0.2\textwidth]{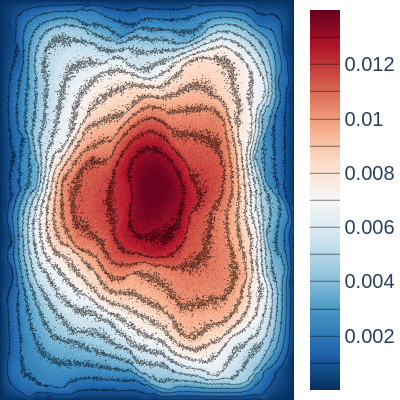}}
    \\
    \subfigure[]{\includegraphics[width=0.2\textwidth]{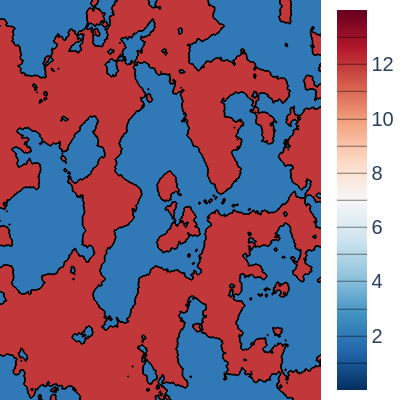}}\hspace{5mm}
    \subfigure[]{\includegraphics[width=0.2\textwidth]{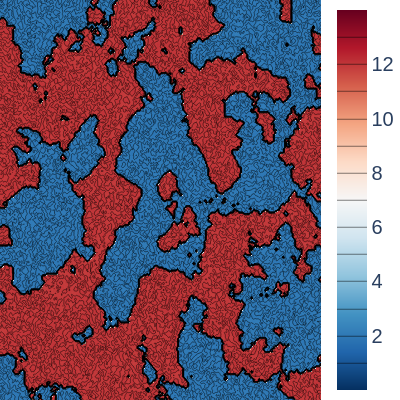}}\hspace{5mm}
    \subfigure[]{\includegraphics[width=0.2\textwidth]{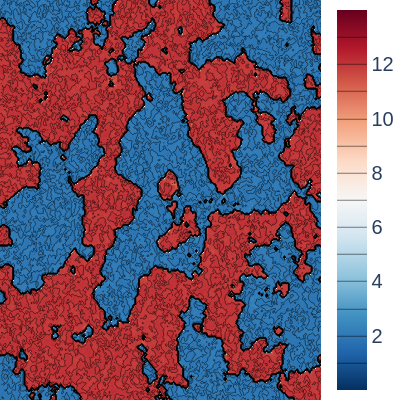}}\hspace{5mm}
    \subfigure[]{\includegraphics[width=0.2\textwidth]{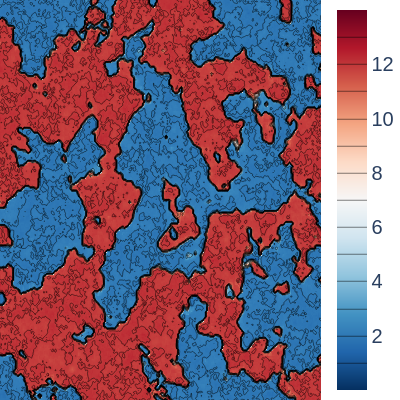}}
    \caption{
        These figures demonstrate the DCNO's prediction of coefficients given solutions with varying noise levels, $\epsilon$:
        (a) Solution, slice at x=0.5.
        (b) Solution in 2D for $\epsilon=0$.
        (c) Solution for $\epsilon=0.01$.
        (d) Solution for $\epsilon=0.1$.
        (e) Coefficient.
        (f) Predicted coefficients for $\epsilon=0$.
        (g) Predicted coefficients for $\epsilon=0.01$.
        (h) Predicted coefficients for $\epsilon=0.1$.
    }
    \label{fig:inversecoefficient}
\end{figure}

In Figure \ref{fig:inversecoefficient}, we display the solution and predicted coefficients for the inverse coefficient identification problem at various levels of noise. Notably, even with a noise level of 10\%, the predicted coefficient successfully recovers the interface present in the ground truth. This resilience to noise highlights the robustness and effectiveness of the coefficient prediction in capturing the underlying structure accurately.

The results of the inverse coefficient identification problem with noise are presented in Table \ref{tab:inverse elliptic noise}. It is worth noting that the memory consumption and training time per epoch remain the same as reported in Table \ref{tab:multi elliptic}. The DCNO model outperforms other methods in this example, which highlights DCNO's ability to effectively address the challenges posed by this ill-posed inverse problem with noisy data. Interestingly, FNO and U-NO, known for their effectiveness in smoothing and filtering high-frequency modes, encounter difficulties in recovering targets that exhibit high-frequency characteristics, such as irregular interfaces, highly oscillatory coefficients, and the presence of Gaussian noise.
In contrast, Dil-ResNet (and MWT to a lesser extent) performs significantly better in this specific problem, particularly when the noise level is higher. The use of dilated convolutions in Dil-ResNet proves advantageous in capturing high-frequency features. While HANO achieves the second-best overall accuracy, it comes at a considerable computational cost.

\begin{table}[H]
    \centering
        \small
    \begin{tabular}{l|cccccc} 
    \toprule
     &\multicolumn{6}{c}{\textbf{Darcy rough}}\\ 
     \textbf{Model} & $\epsilon$=0 & $\epsilon$=0.01 & $\epsilon$=0.03 & $\epsilon$=0.05 & $\epsilon$=0.07 & $\epsilon$=0.1\\   \toprule
    
    \textsc{FNO}  & $0.327$ &$0.328$& $0.330$ & $0.334$ & $0.336$ & $0.339$ \\
    \textsc{MWT}  & $0.103$ & $0.129$ & $0.162$ & $0.183$ & $0.198$ & $0.214$\\
    \textsc{U-NO}  & $0.235$ & $0.236$ & $0.239$ & $0.245$ & $0.251$ & $0.261$\\
    \textsc{GT}  & $0.146$ & $0.172$ & $0.178$ & $0.199$ & $0.214$ & $0.230$\\
    \textsc{HANO} & $0.095$ & $0.118$ & $0.149$ & $0.167$ & \bm{$0.179$} & $0.197$\\
    \textsc{Dil-RESNET} & $0.086$ & $0.122$ & $0.150$ & $0.168$ & $0.185$ & $0.201$\\
    \textsc{DCNO} & \bm{$0.061$} & \bm{$0.115$} & \bm{$0.147$}  & \bm{$0.166$} & $0.180$ & \bm{$0.196$} \\
    \toprule
    \end{tabular}
    \caption{Relative error of the inverse coefficient identification. Also see Figure \ref{fig:inversecoefficient} for solutions and predicted coefficients at various noise levels.}
    \label{tab:inverse elliptic noise}
\end{table}

\subsection{Helmholtz equations}
\label{sec:helmholtz}

\begin{figure}[H]
    \centering
    \subfigure[coefficient $a(x)$]{\includegraphics[width=0.25\textwidth]{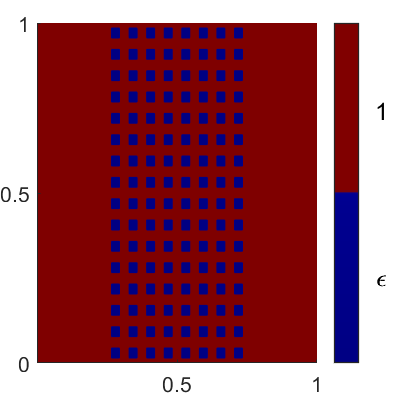}}
    \subfigure[reference solution]{\includegraphics[width=0.25\textwidth]{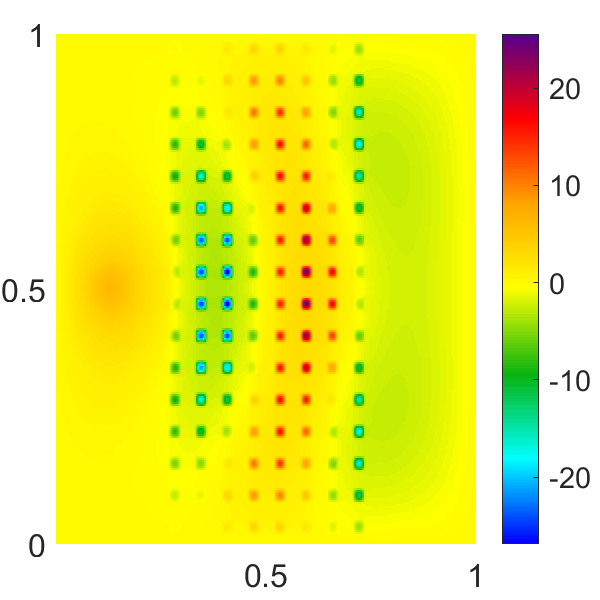}}
    \subfigure[DCNO prediction $\hat{\vu}$]{\includegraphics[width=0.25\textwidth]{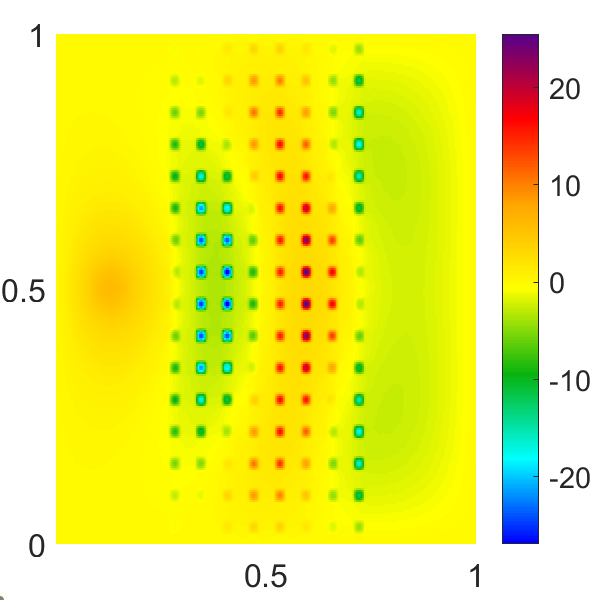}}
    \label{fig:hel}
    \caption{The mapping $a(x) \mapsto \vu$. (a) Heterogeneous coefficient $a(x)$, (b) the reference solution for $\varepsilon^{-1} = 237.3$, which is solved by $\mathcal{P}_1$ FEM implemented in FreeFEM++ \citep{FreeFem}, (c)DCNO predicted solution}
\end{figure}

We evaluate the performance of DCNO for the acoustic Helmholtz equation in highly heterogeneous media, as an illustrative example of multiscale wave phenomena. Solving this equation is particularly challenging and computationally expensive, especially for complex material and large geological models. We adopt the setup described in \citet{freese2021superlocalized},
$$
    \left\{
    \begin{aligned}
    -\mathrm{div}(a(x) \nabla  u(x)) - \kappa^2 u = f(x), & & x \in D,  \\
    u(x)=0, & & x \in \partial D. \\
    \end{aligned}
    \right.
$$
where the coefficient $a(x)$ takes the value 1 or $\varepsilon$ as shown in Figure \ref{fig:hel} with $\varepsilon^{-1} \in \textrm{rand}(128,256)$ , $\kappa = 9$, and
{\footnotesize
$$
    f(x_1, x_2) = 
    \left\{
    \begin{aligned}
    10^4 \exp \Big(\frac{-1}{1-\frac{(x_1-0.125)^2+(x_2-0.5)^2}{0.05^2}} \Big), \quad& 
     (x_1-0.125)^2+(x_2-0.5)^2 < 0.05^2,  \\
    0, \quad &  \textrm{else}.  \\
    \end{aligned}
    \right.
$$    
}
\begin{table}[H]
    \centering
        \small
    \begin{tabular}{l|c|c|c|ccc} 
    \toprule
     & \textbf{Parameters} & \textbf{Memory} & \textbf{Time per}&\multicolumn{3}{c}{\textbf{Relative $L^2$ error}}\\ 
     \textbf{Model} & $\times 10^{6}$     & (GB)              & \textbf{epoch}(s)
    & s=128 & s=256 & s=512\\               
    \toprule
    
    \textsc{FNO}   & $2.37$ & $1.79$ & $5.61$ & $5.033$ & $5.405$  & $6.295$  \\
    \textsc{MWT}   & $9.81$ & $2.54$ & $17.74$ & $2.653$ & $2.731$  & $2.875$  \\
    \textsc{U-NO}   & $16.39$ & $1.57$ & $9.78$ & $2.917$ & $2.852$  & $3.028$  \\
    \textsc{GT}   & $2.22$ & $9.32$ & $35.36$ & $13.505$ & $13.828$  & $16.601$  \\
    \textsc{HANO}   & $8.65$ & $2.33$ & $9.96$ & $\bm{2.472}$ & $3.286$  & $5.189$  \\
    \textsc{DCNO}  & $2.51$ & $3.94$ & $10.14$ & $2.571$ & $\bm{2.682}$  & $\bm{2.748}$  \\
    \toprule
    \end{tabular}
    \caption{Benchmarks on Helmholtz equations at various input resolution $s$. Performance are measured with relative $L^2$ errors ($\times 10^{-2}$), number of parameters, memory consumption for a batch size of 8, and time per epoch for $s=256$ during the training process.}
    \label{tab:hel}
\end{table}

The Helmholtz equation presents a significant challenge due to the highly oscillatory nature of its solution, illustrated in Figure \ref{fig:hel}, and the presence of the $\kappa$-dependent pollution effect. However, the DCNO model has emerged as a notable solution, consistently outperforming other methods in effectively tackling this problem.

\section{Conclusion}

In this paper, we introduce DCNO (Dilated Convolution Neural Operator) as a novel and effective method for learning operators in multiscale PDEs. DCNO leverages the strengths of both Fourier layers, adept at representing low-frequency global components, and convolution layers with multiple dilation rates, capable of capturing high-resolution local details. This hybrid architecture enables DCNO to surpass existing operator methods, providing a highly accurate and computationally efficient approach for learning operators in multiscale settings. Through extensive experiments, we demonstrate the effectiveness of DCNO in addressing multiscale PDEs, highlighting its superior performance and potential for diverse applications.




\bibliographystyle{elsarticle-harv} 
\bibliography{ref}

\end{document}

%% file: math_commands.tex
\usepackage{amsmath,amsfonts,bm}

\def\eqref#1{equation~\ref{#1}}

\def\1{\bm{1}}

\def\va{{\bm{a}}}

\def\vu{{\bm{u}}}

\DeclareMathAlphabet{\mathsfit}{\encodingdefault}{\sfdefault}{m}{sl}
\SetMathAlphabet{\mathsfit}{bold}{\encodingdefault}{\sfdefault}{bx}{n}

